\documentclass[a4paper]{article}

\usepackage{INTERSPEECH2022}
\usepackage{multirow}
\usepackage{comment}

\newcommand{\ind}{\perp\!\!\!\!\perp} 
\usepackage{color, colortbl}
\usepackage{arydshln}
\usepackage{physics}

\title{Better Intermediates Improve CTC Inference}
\name{
Tatsuya Komatsu$^1$, 
Yusuke Fujita$^1$, 
Jaesong Lee$^2$,
Lukas Lee$^2$,
Shinji Watanabe$^3$,
Yusuke Kida$^1$
}
\address{
$^1$LINE Corporation, $^2$NAVER Corporation,\\
$^3$Carnegie Mellon University
}
\email{komatsu.tatsuya@linecorp.com}

\begin{document}

\maketitle
\begin{abstract}
This paper proposes a method for improved CTC inference with searched intermediates and multi-pass conditioning.
The paper first formulates self-conditioned CTC as a probabilistic model with an intermediate prediction as a latent representation and provides a tractable conditioning framework.
We then propose two new conditioning methods based on the new formulation:
(1) Searched intermediate conditioning that refines intermediate predictions with beam-search, 
(2) Multi-pass conditioning that uses predictions of previous inference for conditioning the next inference.
These new approaches enable better conditioning than the original self-conditioned CTC during inference and improve the final performance.
Experiments with the LibriSpeech dataset show relative 3\%/12\% performance improvement at the maximum in test clean/other sets compared to the original self-conditioned CTC.
\end{abstract}

\section{Introduction}
End-to-end automatic speech recognition (ASR) directly maps input speech to text,
realizing a simple ASR pipeline by eliminating the conventionally required hand-crafted pronunciation dictionary.
Popular end-to-end ASR approaches are autoregressive models, 
such attention-based encoder-decoders~\cite{Chorowski15_NIPS} and recurrent neural network transducers~\cite{Graves12_ICMLRLW}, 
in which autoregressive decoders predict a token given previously estimated tokens with a sequence of encoded audio features.
These autoregressive ASR methods have shown state-of-the-art performance, 
with powerful neural architectures such as Transformer \cite{Dong18_ICASSP,karita2019comparative} and Conformer \cite{gulati20_interspeech,Pengcheng21_icassp}.
Non-autoregressive ASR methods, as well, have been attracting interest because of their ability to perform fast inference that can predict all tokens simultaneously.
Non-autoregressive ASR can be broadly divided into two types: 
methods based on non-autoregressive decoders that iteratively improves the decoded results~\cite{chan2020imputer,Chen21_SPL,Higuchi20b_interspeech,Chi21_NAACL} 
and connectionist temporal classification (CTC)~\cite{Graves06_icml,amodei2016deep} based methods.
The CTC-based models seek to obtain a better representation by the audio encoder,
using multi-task learning with hierarchical targets~\cite{toshniwal2017multitask,ramon2018hierarchical,krishna2018hierarchical} and additional auxiliary losses based on additional modules~\cite{tjandra2020deja}.
In particular, interCTC and self-conditioned CTC~\cite{lee21_icassp, nozaki21_interspeech} have reported performance comparable to conventional autoregressive encoder-decoder models without language models in several benchmarks \cite{higuchi2021comparative}, although the decoding itself is a CTC greedy decoding (best path search).

InterCTC~\cite{lee21_icassp} and self-conditioned CTC~\cite{nozaki21_interspeech} consist of a transformer/conformer-based audio encoder and a CTC decoder.
The greedy decoding can be viewed as frame-by-frame classification of the text and has the disadvantage that the output text at each time step can only be predicted independently under conditional independence. 
A key feature of interCTC and self-conditioned CTC is how to capture textual information using only an audio encoder.
InterCTC makes predictions of the output sequence in the intermediate layers and takes intermediate CTC loss as well as the final layer.
Self-conditioned CTC extends interCTC to explicitly utilize intermediate predictions for conditioning the subsequent layers by adding the intermediate prediction to the input of the next layer.
A recent comparative study~\cite{higuchi2021comparative} reported that self-conditioned CTC showed the best performance among the latest non-autoregressive models, without any special decoder for the output tokens.
Thus, the conditioning framework using intermediate predictions has proven to be very powerful. 
However, the mechanism behind the conditioning has not been fully explained.
Formulating this mechanism in a tractable form, such as a probabilistic model, allows for more advanced discussions and extensions.

In the sequence-to-sequence modeling, there are several works that formulate the probability model of the output as a marginalized distribution with a latent representation~\cite{van2018parallel,gu2018nonautoregressive,lee2018deterministic,shu2020latent}.
In particular, recent research~\cite{dalmia2021searchable} has decomposed the entire task into its subtasks, 
and directly refining the latent representation of the subtasks to improve the overall performance.
Self-conditioning, which explicitly handles intermediate prediction, could be similarly formulated as a probability model with intermediate prediction as a latent representation.

This paper reformulates the original self-conditioned CTC with an approximate probabilistic model.
With this formulation, we introduce two other conditioning methods to inject external knowledge for improved inference: (1) conditioning using beam-searched intermediate outputs, (2) multi-path conditioning, in which the prediction results obtained from previous inference are used to condition the next inference. 
We confirm experimentally that the inference of the proposed method based on the new formulation yields further improvement on the strong self-conditioned CTC baseline.
\section{Self-conditioned CTC}
\subsection{CTC-based ASR}
\label{sec:conformerctc}
Let $Y=(y_l \in \mathcal{V} \mid l=1,\dots,L)$ be an $L$-length label sequence with vocabulary $\mathcal{V}$ and $X=(\mathbf{x}_t \in \mathbb{R}^D \mid t=1,\dots,T)$ be a $T$-length audio feature sequence with feature dimension size $D$.
End-to-end ASR can be formulated as a search problem of finding the most probable output sequence $\hat{Y}$ as follows:
\begin{align}\label{eq:prop}
    \hat{Y} = \underset{Y}{\mathsf{argmax}}~p(Y \mid X). 
\end{align}
CTC models this $X \rightarrow Y$ mapping with
a label probability for each time step and its alignment representing a text sequence.

\noindent\textbf{Encoding of audio sequence:} Let us consider a CTC-based system based on $N$-layer Conformer encoders.
First, an audio sequence $X(=X^{(0)})$ is fed into the Conformer encoders and is converted to the $N$-th feature sequence $X^{(N)}$.
Here, the input and output of each layer are represented as follows,
\begin{align}\label{eq:encoder}
    X^{(n)} = \mathsf{Encoder}^{(n)}(X^{(n-1)}).
\end{align}
Then, the vector $\mathbf{x}_t^{(N)}$ at time step $t$ of $X^{(N)}$ is mapped to a probability distribution on the vocabulary labels and an extra $\mathsf{blank}$ symbol as $\mathcal{V}'= \mathcal{V} \cup \{\mathsf{blank}\}$, where the $\mathsf{blank}$ represents the repetition of the preceding token.
This mapping is performed with a linear transformation and a softmax function as, 
\begin{align}\label{eq:softmax}
    Z = \mathsf{Softmax}(\mathsf{Linear}_{D\rightarrow |\mathcal{V}'|}(X^{(N)})),
\end{align}
where $Z = (\mathbf{z}_t \in (0,1)^{|\mathcal{V}'|} \mid t=1,\dots,T)$ is a sequence of the label probability at each time step. The element $z_{t,k}$ is interpreted as the posterior probability of $k$-th label at time $t$.
Note that we denote $\mathsf{Linear}(\cdot)$ and
$\mathsf{Softmax}(\cdot)$ as operations to each vector in a sequence.

\noindent\textbf{CTC decoding:}
Since $Z$ is a sequence with the encoder input length, 
we need to align it to the text-domain sequence.
CTC introduces an alignment path $A=\left(a_t \in \mathcal{V}' \mid t=1,\dots,T\right)$ and the collapsing function $\mathcal{B}(A)$ that removes all repeated labels and blank symbols in the alignment.
The alignment $a_t$ represents the label at time step $t$.
$a_t$ follows the obtained label distribution $a_t\sim\mathbf{z}_t$, 
and since label estimation is performed each time step, there is conditional independence $(a_t \ind a_{\ne t} \mid X)$.
Thus, the probability of alignment $p(A\mid X)$ is obtained as the product of the label probabilities over $T$ time steps,
\begin{align}\label{eq:alignment}
    p(A=(k_1,...,k_T) \mid X) &= \prod_t p(a_t=k_t \mid X) \\
    & = \prod_t z_{t,k_t}.
\end{align}
Using this probability distribution over the alignment $A$, the probability $p(Y\mid X)$ in Eq.~\ref{eq:prop} is expressed as the sum of the probabilities of all possible alignment paths of $Y$, i.e.,
\begin{align}\label{eq:pathprob}
    p(Y \mid X) = \sum_{A\in\mathcal{B}^{-1}(Y)}p(A\mid X).
\end{align}

To train neural network parameters using CTC, the loss is defined as the negative log-likelihood of $Z$ for all possible paths corresponding to the target sequence $Y^{(\mathsf{tgt})}$,
\begin{align}\label{eq:lossctc}
    \mathcal{L}_\mathsf{ctc}(Z, Y^{(\mathsf{tgt})}) = - \log \sum_{A\in\mathcal{B}^{-1}(Y^{(\mathsf{tgt})})}\prod_{t} p(a_t\mid X).
\end{align}
For inference, we need to find the most probable label sequence and its alignment based on the label probabilities at all time steps. 
However, this computational complexity grows exponentially with the sequence length, making it impractical.
For the efficient search of the best alignment, prefix beam search \cite{maas2014FirstPassLV} can be used with a language model.
Joint CTC/attention decoding \cite{watanabe2017CTCAttention} is also a popular method using an attention-based decoder.
When neither decoder nor language model is available, the best path search is a commonly used approximation.
It assumes that the most probable label sequence $Y^*$ and its alignment $A^*$ corresponds to:
\begin{align}\label{eq:bestpath}
    Y^* &\approx \mathcal{B}(A^*) \\
    \text{where}~~a_t^* &= \underset{a_t}{\arg\max}~p(a_t\mid X).
\end{align}
It is not guaranteed that the best label sequence will be found, but it is empirically confirmed that in most cases the solution is sufficient~\cite{higuchi2021comparative}.

\subsection{Conditioning with intermediate CTC}
\label{sec:interctc}
InterCTC \cite{lee21_icassp} introduces additional CTC predictions from intermediate encoder blocks.
An intermediate label prediction for the $n$-th encoder block $Z^{(n)} = (\mathbf{z}^{(n)}_t \in (0,1)^{|\mathcal{V}'|}| t=1,\dots,T)$ is computed similar to Eq. \ref{eq:softmax} as:
\begin{align}\label{eq:interpred}
    Z^{(n)} = \mathsf{Softmax}(\mathsf{Linear}_{D\rightarrow |\mathcal{V}'|}(X^{(n)})).
\end{align}
The losses for the intermediate predictions are computed as well as the original CTC loss (Eq. \ref{eq:lossctc}), and the total loss is a weighted sum of the original and the intermediate CTC losses:
\begin{equation}\label{eq:interctc}
    \mathcal{L}_\mathsf{ic} = (1-\lambda)\mathcal{L}_\mathsf{ctc}(Z,Y) + \frac{\lambda}{|\mathcal{N}|}\sum_{n \in \mathcal{N}} \mathcal{L}_\mathsf{ctc}(Z^{(n)},Y),
\end{equation}
where $\lambda \in (0,1)$ is a mixing weight and $\mathcal{N}$ is a set of layer indices for intermediate loss computation.

Self-conditioned CTC \cite{nozaki21_interspeech} further utilizes the intermediate prediction for conditioning the subsequent encoders.
The intermediate prediction $Z^{(n)}$ in Eq.~\ref{eq:interpred} is projected back to $D$-dimensional audio representation as 
\begin{align}\label{eq:H}
    H^{(n)} = \mathsf{Linear}_{|\mathcal{V}'|\rightarrow D}(Z^{(n)}),
\end{align}
and added to the input of the next encoder:
\begin{align}\label{eq:selfcond}
    X'^{(n)} &= X^{(n)} + H^{(n)}, 
\end{align}
where $n \in \mathcal{N}$ and $X'^{(n)}$ is a new input for the next encoder.
$\mathsf{Linear}_{|\mathcal{V}'|\rightarrow D}(\cdot)$ maps a $|\mathcal{V}'|$-dimensional vector into a $D$-dimensional vector for each element in the input sequence.

\section{Proposed method}
In this section, we provide a probabilistic formulation for self-conditioned CTC to describe its behavior.
We then propose a new formulation with best path approximation and introduce improved inference based on two new types of conditioning.
\subsection{Probabilistic formulation of self-conditioning}
Self-conditioned CTC achieves improved encoding by conditioning the audio encoder on intermediate predictions $Y^{\mathsf{inter}}$.
Eq.~\ref{eq:prop} can be formulated as a marginal distribution with latent representation as prior works~\cite{van2018parallel,gu2018nonautoregressive,lee2018deterministic,shu2020latent},
\begin{align}\label{eq:marginal}
    \hat{Y} &= \underset{Y}{\mathsf{argmax}}~\sum_{Y^\mathsf{inter}}p(Y, Y^\mathsf{inter} \mid X) \\
            &= \underset{Y}{\mathsf{argmax}}~\sum_{Y^\mathsf{inter}}p(Y \mid Y^\mathsf{inter}, X)p(Y^\mathsf{inter} \mid X)
\end{align}
Self-conditioned CTC feeds hidden representation $H^{\mathsf{inter}}$ back to the encoder as Eq~\ref{eq:selfcond}.
$H^{\mathsf{inter}}$ is the linear projection of the posterior distribution $Z^{\mathsf{inter}}$.
Here, describing the linear transformation of Eq.~\ref{eq:H} as $\mathbf{W}\in\mathbb{R}^{D\times|\mathcal{V}'|}$, 
the vector at each time step of $H^{\mathsf{inter}}$ can be written as follows,
\begin{align}\label{eq:hidden}
    \mathbf{h}^\mathsf{inter}_{t} &= \mathbf{W} \mathbf{z}_{t} \\
     &= \sum_k \mathbf{w}_k z_{t,k}.
\end{align}
Let the vector $\mathbf{w}_k\in\mathbb{R}^d$ of $\mathbf{W}$ be the mapping $g(k)$ that maps the $k$-th label to the $D$-dimensional feature, 
and since $z_{t,k}$ is the label probability distribution of the alignment $a_{t,k}$, 
Eq.~(16) can be described as conditional expectation as follows:
\begin{align}\label{eq:expect}
     \sum_k \mathbf{w}_k z_{t,k} &= \sum_k g(a_{t,k}) p(a_{t,k}\mid X) \\
     &= \mathbf{E} \left[g(a_{t,k}) \mid X\right] ,
\end{align}
where $\mathbf{E}[\cdot]$ denotes the expectation. 
For an entire sequence, since $a_{t,k}$ at each time step is conditionally independent, we obtain
\begin{align}\label{eq:expect_align}
    H^\mathsf{inter} &=  \mathbf{E} \left[g(A) \mid X\right] \\
    &=  \mathbf{E} \left[g\left(\mathcal{B}^{-1}(Y)\right) \mid X\right].
\end{align}
Thus, $H^\mathsf{inter}$ can be interpreted as the conditional expectation of $Y$ given $X$.
Therefore, self-conditioned CTC, which uses $H^\mathsf{inter}$ for conditioning, makes the following approximation of Eq.~14:
\begin{align}\label{eq:expect_selfcond}
    \sum_{Y^\mathsf{inter}}p(Y \mid Y^\mathsf{inter}, X)p(Y^\mathsf{inter} \mid X) \approx p(Y \mid \mathbf{E}[Y^\mathsf{inter}],   X),
\end{align}
where the marginalization with $Y$ is replaced by the conditioning on expectation with $Y$.
This approximation depends on the `confidence' of the probability $p(Y^{\mathsf{inter}}\mid X)$, and especially predictions at lower layers are often inadequate.
This paper focuses and seeks better approximation of self-conditioning.

\subsection{New formulation with best path approximation}\label{sec:bestpath}
This paper provides a new formulation of self-conditioned CTC by a different approximation to Eq.~\ref{eq:marginal},
that replaces the marginalization by maximization:
\begin{align}
    \hat{Y} &= \underset{Y}{\mathsf{argmax}}~\sum_{Y^\mathsf{inter}}p(Y \mid Y^\mathsf{inter}, X)p(Y^\mathsf{inter} \mid X)\nonumber \\
            & \approx \underset{Y}{\mathsf{argmax}}~p(Y \mid \tilde{Y}^\mathsf{inter},  X), \label{eq:new_cond}\\
    \textrm{where}~~~&\tilde{Y}^\mathsf{inter} =\mathsf{\mathsf{argmax}}_{Y^\mathsf{inter}}~p(Y^\mathsf{inter} \mid X).
\end{align}
This method replaces the intractable marginalization by deterministically conditioning with the most probable single prediction, known as the Viterbi approximation.

To realize self-conditioned CTC by this approximation, Eq.~\ref{eq:H} also requires an another form.
First, we estimate the most probable label sequence $\tilde{Y}^\mathsf{inter}$ based on the intermediate label prediction, $Z$.
As described in Section~\ref{sec:conformerctc}, we can use search techniques such as prefix beam search~\cite{maas2014FirstPassLV} with a language model, but we first formulate it with the best path approximation as in Eq.~\ref{eq:bestpath}:
\begin{align}\label{eq:besty}
    \tilde{Y}^\mathsf{inter} &\approx \mathcal{B}(A^*) \\
    \text{where}~~a_t^* &= \underset{a_t}{\mathsf{argmax}}~p(a_t\mid X). \label{eq:besty2}
\end{align}
The resulting best path sequence $A^*$ is transformed into a representation $H$ by embedding it in $D$-dimensional feature:
\begin{align}
    H=\mathsf{Embedding}_{(1\rightarrow D)} (A^*)
\end{align}
This operation $\mathsf{Embedding}$ projects the label of each time step $a_t$ of $A$ into the $D$-dimensional feature space, essentially the same as $g(\cdot)$ in Eq.~\ref{eq:expect}.
The resulting $H$ is added to the next input as Eq.~\ref{eq:selfcond} to condition subsequent encoders.

The advantage of this approximation is that the conditioning used here is a deterministically obtained label sequence $a_t ^{\ast}$ in Eq. \ref{eq:besty2}, rather than a sequence of probabilities or expectations $\mathbf{z}_t$, as introduced in Section \ref{sec:interctc}.
Therefore, to obtain $a_t ^{\ast}$ during inference, it makes it possible to apply sequence-level techniques like beam search, 
including combinations of external language models, to refine the information for conditioning $\tilde{Y}^\mathsf{inter}$ in Eq.~\ref{eq:new_cond}.
Accompanying this new approximation, we propose two new conditioning approach for inference.

\subsection{Proposed inference with new conditioning approaches}
\subsubsection{Searched intermediate conditioning.}\label{sec:searched}
The best path approximation allows the text-domain sequence to be used for conditioning.
Therefore, it is possible to obtain better intermediate prediction by using text-domain techniques, e.g., beam search with external language models:
\begin{align}
    Y^{\mathsf{search}} = \mathsf{Beam}\left(p\left(Y_l^{\mathsf{inter}} \mid X, Y_{1:l-1}^{\mathsf{inter}}\right)\right),
\end{align}
which is the prediction obtained by autoregressive LM, in contrast to the conditionally independent prediction in Eq.~\ref{eq:besty}.
But $Y^{\mathsf{search}}$ is the sequence of the text domain, which requires conversion to a time step alignment for conditioning.
In this paper, we obtain the alignment of $Y$ as the most probable path to $Z$ by using Viterbi algorithm:
\begin{align}\label{eq:search}
    A^{\mathsf{search}} = \textsf{Viterbi}(Y^{\mathsf{search}}, Z)
\end{align}
The resulting $A^{\mathsf{search}}$ is converted to the representation $H$ for conditioning using Eq.~\ref{eq:selfcond}.

\subsubsection{Multi-pass conditioning}\label{sec:multipass}
Another new conditioning method is multi-pass conditioning.
This method is based on multiple inferences and uses the predicted results obtained in the previous inference as information for conditioning at the next inference.
In each inference, the input features remain unchanged,
and only the representation for conditioning $H$ is replaced by the results obtained in the previous inference.
Let $\hat{Y}^{(m)}$ denotes the final result obtained in the $m$-th inference, 
then $\hat{Y}^{(m)}$ is used in for condition in the $(m+1)$-th inference.
At each conditioning layer, $\hat{Y}^{(m)}$ is aligned with intermediate prediction $Z^{(n)}$ and the obtained alignment is converted to $H^{(n)}$, using $\mathsf{Viterbi}(\cdot)$ and $\mathsf{Embedding}(\cdot)$ as Eq.~\ref{eq:search}.
The result of the inference is naturally better than the prediction of the intermediate layers. 
Therefore, the condition based on that better result can be used to improve the next inference.

\begin{table*}[ht]
\centering
\caption{
Experimental results using greedy search and LM+beamsearch for the final outputs.
}\vspace{-3mm}
\label{tab:result}
\resizebox{\textwidth}{!}{%
\begin{tabular}{lccccccccc}
\hline
\multirow{2}{*}{model} & \multirow{2}{*}{output decoding} & \multicolumn{4}{c}{LibriSpeech   (100h)} & \multicolumn{4}{c}{LibriSpeech   (960h)} \\ 
 &  & dev-clean & dev-other & test-clean & test-other & dev-clean & dev-other & test-clean & test-other \\ \hline
\multicolumn{10}{l}{\textit{decoding  with CTC greedy search}} \\
CTC & greedy & 10.21 & 24.19 & 10.65 & 25.00 & 3.79 & 9.76 & 3.93 & 9.88 \\
InterCTC & greedy & 7.76 & 20.59 & 8.16 & 21.21 & 3.09 & 8.49 & 3.30 & 8.36 \\
SelfCond & greedy & 7.51 & 20.04 & 7.86 & 20.65 & 2.86 & 7.87 & 3.06 & 7.91 \\
\hdashline
{SelfCond + best path} & greedy & 7.20 & 19.81 & 7.78 & 20.76 & 2.75 & 7.88 & 3.06 & 7.83 \\
~~+ searched cond.$^\dagger$ & greedy & \textbf{6.45} & \textbf{18.45} & \textbf{7.11} & \textbf{18.57} & \textbf{2.68} & \textbf{6.95} & \textbf{2.98} & \textbf{7.07} \\
~~+ 2-pass & greedy & 6.95 & 19.25 & 7.51 & 20.23 & 2.71 & 7.23 & 3.01 & 7.32 \\ \hline\hline
\multicolumn{2}{l}{\textit{decoding  with Transformer LM + search}} &  &  &  &  &  &  &  &  \\
SelfCond & beam search & 5.05 & 15.44 & 5.40 & 15.61 & 2.19 & 5.96 & 2.45 & 6.08 \\
\hdashline 
{SelfCond + best path} & beam search & 4.88 & 14.88 & 5.40 & 15.53 & 2.12 & 5.81 & 2.48 & 5.96 \\
~~+ searched cond.$^\dagger$ & beam search & \textbf{4.60} & \textbf{14.00} & \textbf{5.07} & \textbf{14.47} & 2.17 & 5.66 & 2.48 & 5.97 \\
~~+ 2-pass cond. & beam search & 4.74 & 14.14 & 5.19 & 14.69 & \textbf{2.07} & \textbf{5.53} & \textbf{2.44} & \textbf{5.77} \\ \hline
\vspace{-3mm}
\end{tabular}%
}\\{\footnotesize $^\dagger$ For searched conditioning approache, we applied Transformer LM + beam search on the intermediate predictions.}
\vspace{-3mm}
\end{table*}

\section{Experiment}
To verify the effectiveness of the proposed method, we conducted experiments 
by using ESPnet \cite{watanabe18_interspeech, Pengcheng21_icassp} with almost the same hyperparameters.

\subsection{Data}
The experiments are carried out using the LibriSpeech~\cite{panayotov2015librispeech} dataset, utterances from read English audio books. 
We trained the models using the 100-hour subset (LS100) or the 960-hour full set (LS960), and we used the standard development and test sets for evaluation. 
For the input feature, 80-dimensional Mel-scale filterbank coefficients with three-dimensional pitch features were extracted using Kaldi toolkit \cite{Povey11_ASRU}.
Speed perturbation \cite{Ko15_interspeech} and SpecAugment \cite{Park19_interspeech} were also applied to the training data.

\subsection{Model configurations}
\textbf{CTC:}
We used the Conformer-CTC model as described in Section \ref{sec:conformerctc}.
The number of layers $N$ was 18, and the encoder dimension $D$ was 256.
The convolution kernel size and the number of attention heads were 15 and 4, respectively.
The feed-forward layer dimension in the Conformer blocks was set to 1024.
The model was trained for 50 epochs, and the final model was obtained by averaging model parameters over 10-best checkpoints in terms of validation loss values.
The batch-size is set to 1024 for LS960 and 128 for LS100.

\noindent\textbf{InterCTC and +SelfCond:}
For intermediate CTC and self-conditioned CTC (Section~\ref{sec:interctc}), 
we applied five intermediate CTC predictions at the set of layer indices $\mathcal{N} = \{3,6,9,12,15\}$ with $\lambda = 0.5$ in Eq.~\ref{eq:interctc},  conditioning with the intermediate CTC predictions was applied according to Eq.~\ref{eq:selfcond}.
Other configurations are identical to the baseline CTC.

\begin{figure}[t]
  \centering
  \includegraphics[width=0.8\linewidth]{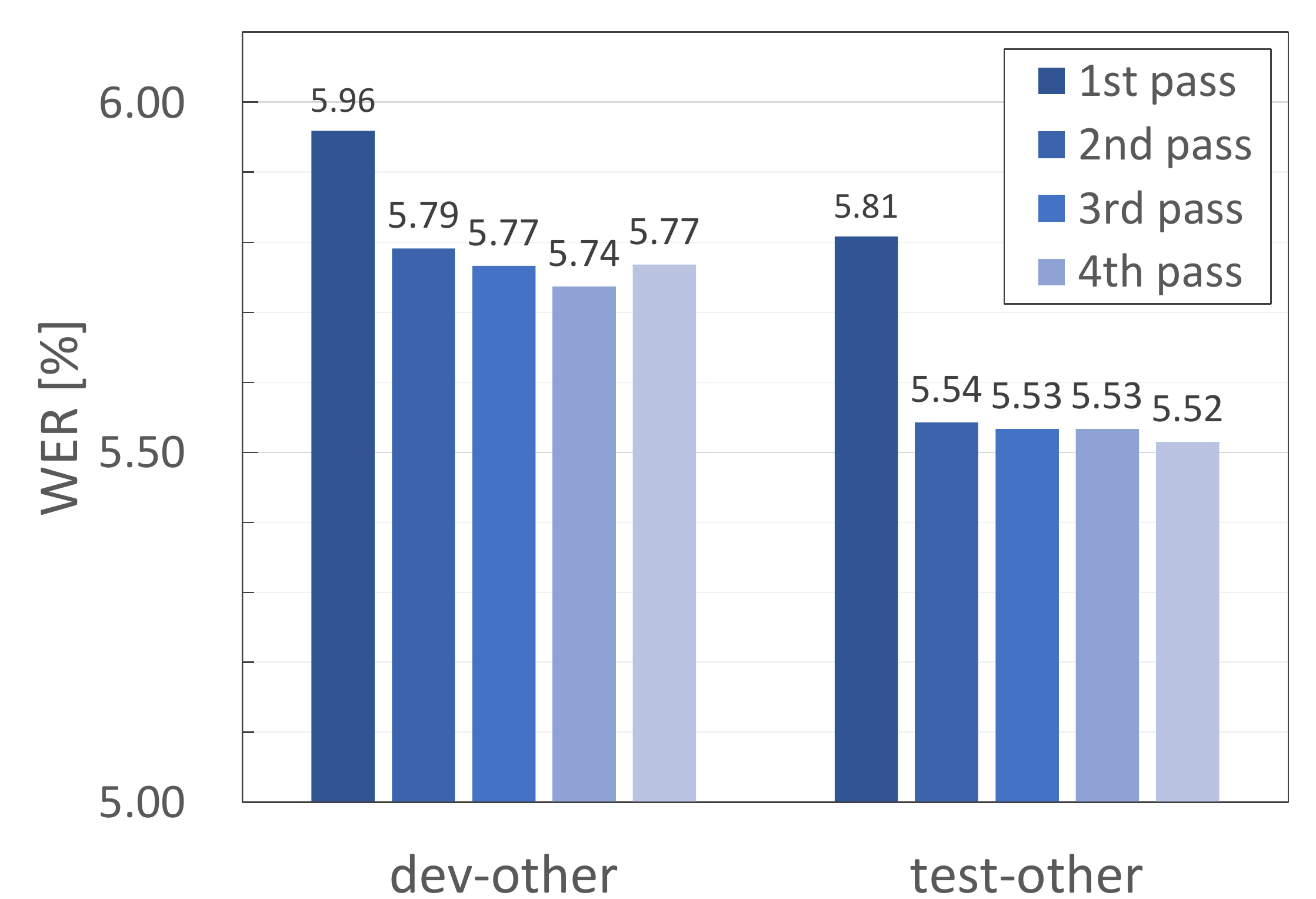}\vspace{-2mm}
  \caption{
  Multi-pass results of LS960. The 1st pass is normal intermediate conditioning; 2nd and subsequent passes are conditioned on results of previous pass. The final output of each pass are obtained with beam search and the Transformer LM.}
  \vspace{-4mm}
  \label{fig:multipass}
\end{figure}

\begin{table}[t]
\centering
\caption{Comparison of LMs used for searcher condition with LS960. Since the final output includes the performance of the LM+beam search, the results of the CTC greedy search are also shown for comparison of the encoder output.}\vspace{-3mm}
\label{tab:lms}
\begin{tabular}{cccccc}
\hline
\multirow{2}{*}{condition} & \multirow{2}{*}{decode} & \multicolumn{2}{c}{dev} & \multicolumn{2}{c}{test} \\
 &  & clean & other & clean & other \\\hline
4-gram & beam & 2.39 & 6.50 & 2.67 & 6.54 \\
Transformer & beam & 2.17 & 5.66 & 2.48 & 5.97 \\\hline\hline
4-gram & greedy & 2.77 & 7.51 & 3.02 & 7.29 \\
Transformer & greedy & 2.68 & 6.95 & 2.98 & 7.07\\\hdashline
Oracle & greedy & 2.24 & 5.18 & 2.49 & 5.42 \\\hline
\end{tabular}\vspace{-6mm}
\end{table}

\noindent\textbf{SelfCond + best path approx.:}
All configuration of the proposed model is identical with the SelfCond baseline except for the conditioning method as described in Section~\ref{sec:bestpath}.
For beam search of intermediate and final results, we used two different LMs:
Transformer-based pretrained LM provided by ESPnet, and 4-gram LM trained with LibriSpeech 960h full set.

\subsection{Results}
Table~\ref{tab:result} summarizes the experimental results.
First, it can be seen that new self-conditioned CTC with the best path approximation boosts the performance of the strong self-conditioned CTC beseline.
This can be explained because the conventional self-conditioned CTC uses expectation of the intermediate prediction that includes the ambiguity of the prediction.
Of course, conditioning with deterministic predictions might have negative impacts if the predictions are wrong, 
but under this experimental settings, there were only positive effects.

Both conditioning by searched intermediates and multi-pass conditioning are shown to further improve the final results.
In LS960, the proposed method shows superior results for all data sets, although the amount of improvement appears to be small because the results were already good before using the proposed conditioning.
For the smaller LS100, the improvement by the proposed method is more significant.

\noindent\textbf{Use of other language model}
Table~\ref{tab:lms} compares the results of using different LMs for searched conditioning (Section~\ref{sec:searched}).
It is clear that the more powerful LMs give better results.
The greedy decoding of the final result can be seen as a refinement of embedding itself.
It can be seen that performing conditioning with better information will also yield better final results.
This table also shows the results of oracle conditioning, which is conditioning based on ground truth. This is the best performance that can be achieved with this architecture. 

\noindent\textbf{Multi-pass decoding}
Investigation of the impact of multi-pass conditioning (Section~\ref{sec:multipass}) are shown 
in Figure~\ref{fig:multipass}.
The second pass shows the most WER improvement, and the third and later passes seem to be mostly saturated.
The clean set shows the same trend as other set, although the improvement is slight because the 1st pass results are already high.

\section{Conclusions}
In this paper, we formulated self-conditioned CTC as a probabilistic model of output sequence marginalized with intermediate prediction for more tractable conditioning.
On this formulation, we also introduced new conditioning approaches for improved inference; searched intermediate conditioning and multi-pass conditioning.
Experimental results showed that our new formulation and conditioning improved the inference.

\bibliographystyle{IEEEtran}
\bibliography{mybib}

\end{document}